\documentclass{article}

\usepackage{arxiv}

\usepackage[utf8]{inputenc} 
\usepackage[T1]{fontenc}    
\usepackage{hyperref}       
\usepackage{url}            
\usepackage{booktabs}       
\usepackage{amsfonts}       
\usepackage{nicefrac}       
\usepackage{microtype}      
\usepackage{lipsum}		
\usepackage{graphicx}
\usepackage{natbib}
\usepackage{doi}

\usepackage{adjustbox}
\usepackage{amsmath}
\usepackage{float}

\usepackage{fontawesome}
\usepackage{dblfloatfix}

\title{ViG-LRGC: Vision Graph Neural Networks with Learnable Reparameterized Graph Construction}

\author{
{\hspace{1mm}\href{https://orcid.org/0009-0002-8510-2884}{Ismael Elsharkawi}} \\
	The American University in Cairo\\
	\texttt{ismaelelsharkawi@aucegypt.edu} \\
	\And
    {\hspace{1mm}\href{https://orcid.org/0000-0003-0042-9790}{Hossam Sharara}} \\
	The American University in Cairo\\
	\texttt{hossam.sharara@aucegypt.edu} \\
	\And
    {\hspace{1mm}\href{https://orcid.org/0000-0001-8109-1845}{Ahmed Rafea}} \\
	The American University in Cairo\\
	\texttt{rafea@aucegypt.edu} \\
}

\date{}

\renewcommand{\shorttitle}{\textit{ViG-LRGC}}

\hypersetup{
pdftitle={A template for the arxiv style},
pdfsubject={q-bio.NC, q-bio.QM},
pdfauthor={David S.~Hippocampus, Elias D.~Striatum},
pdfkeywords={First keyword, Second keyword, More},
}

\begin{document}
\maketitle

\begin{abstract}
Image Representation Learning is an important problem in Computer Vision. Traditionally, images were processed as grids, using Convolutional Neural Networks or as a sequence of visual tokens, using Vision Transformers. Recently, Vision Graph Neural Networks (ViG) have proposed the treatment of images as a graph of nodes; which provides a more intuitive image representation. The challenge is to construct a graph of nodes in each layer that best represents the relations between nodes and does not need a hyper-parameter search. ViG models in the literature depend on non-parameterized and non-learnable statistical methods that operate on the latent features of nodes to create a graph. This might not select the best neighborhood for each node. Starting from $\textit{k}$-NN graph construction to HyperGraph Construction and Similarity-Thresholded graph construction, these methods lack the ability to provide a learnable hyper-parameter-free graph construction method. To overcome those challenges, we present the Learnable Reparameterized Graph Construction (LRGC) for Vision Graph Neural Networks. LRGC applies key-query attention between every pair of nodes; then uses soft-threshold reparameterization for edge selection, which allows the use of a differentiable mathematical model for training. Using learnable parameters to select the neighborhood removes the bias that is induced by any clustering or thresholding methods previously introduced in the literature. In addition, LRGC allows tuning the threshold in each layer to the training data since the thresholds are learnable through training and are not provided as hyper-parameters to the model. We demonstrate that the proposed ViG-LRGC approach outperforms state-of-the-art ViG models of similar sizes on the ImageNet-1k benchmark dataset. 
\end{abstract}

\section{Introduction}
\label{sec:intro}
The task of pretraining backbones in Computer Vision is an important one that has been tackled extensively in the literature. The purpose of pre-training a model is to have a backbone that could be fine-tuned later to another downstream task. Traditionally, Computer Vision backbones are pre-trained on the image classification task using the following datasets: ImageNet-1K \citep{ImageNet}, ImageNet-21K \citep{ridnik2021imagenet21kpretrainingmasses}, or JFT-300M \citep{sun2017revisitingunreasonableeffectivenessdata} (which is not publicly available). Concerning the type of backbone, there are four of them in the literature, which differ in the way they represent an image:  Convolutional Neural Networks (CNN) \citep{726791}, Vision Transformers \citep{dosovitskiy2020vit}, Multi-Layer Perceptron Mixers (MLP Mixers) \citep{tolstikhin2021mixer}, and Vision Graph Neural Networks (ViG) \citep{han2022vig}.

CNNs process an image as a grid of pixels, where learnable moving filters are applied to an image. The number of filters increases as the network goes deeper. By using learnable filters, spatial locality and shift invariance are provided. CNNs became the common type of Deep Learning Network in Computer Vision after AlexNet \citep{NIPS2012_c399862d}, which was trained on ImageNet-1k.

Transformers \citep{vaswani2023attentionneed} were introduced in NLP to process sequences of tokens, where \textit{self-attention} is applied between tokens. The dot product of the key and query vector is a main concept introduced in self-attention, which represents the importance of the token corresponding to the key with respect to the token corresponding to the query. Likewise, Vision Transformers \citep{dosovitskiy2020vit} apply self-attention between patches of an image (the visual tokens). Nevertheless, applying self-attention comes at the cost of a large number of FLOPs and parameters. 

Multi-Layer Perceptron Mixers (MLP-Mixers) introduced the concept of channel mixing and patch mixing. Using MLP-Mixers, an image would be treated as a table of $patches \times channels$, where each \textit{mixing} operation is a transpose operation, followed by a linear projection on the \textit{channel} or the \textit{patch} dimension, and, hence the name mixing.

More recently, Vision Graph Neural Networks \citep{han2022vig} represent an image as a graph of nodes, where each node would represent a patch in the input image. This method of image representation is more flexible than CNNs since a node neighborhood has no restriction, unlike the local neighborhood of a pixel in a grid. Added to that, in ViG, nodes have a relatively smaller neighborhood, unlike ViT, which performs self-attention between every possible pair of tokens.

ViG was the first attempt to use Graph ML to pre-train a Computer Vision model. However, Graph ML is usually used to process natively represented graph data, such as citation networks \citep{sen2008collective}, social networks \citep{sage} and biochemical graphs \citep{wale2008comparison}. More recently, 3D Computer Vision tasks started getting tackled with Graph ML. These include point cloud classification and segmentation. The challenge was to construct a graph that best represented the point cloud in addition to formulating the best possible GNN layer architecture. For example, DGCNN \citep{DBLP:journals/corr/abs-1801-07829} used a \textit{k}-Nearest Neighbor (\textit{k}-NN) graph construction methodology and introduced EdgeConv, which is GNN layer type. Then, to overcome the oversmoothing phenomenon, DeepGCNs \citep{DeepGCNs} used a linear layer before and after the GNN. Oversmoothing means that the latent representations of nodes become very similar to one another after multiple aggregation operations. DeepGCNs also proposed the use of dilated k-NN graph construction, which means that the $k$ neighbors are going to be chosen in a stochastic manner from the $k\times d$ nearest neighbors for every node, such that $d$ is the dilation factor. This is important to mention since ViG used the same methodology of DeepGCNs; however, nodes represent patches of an image rather than point clouds.

To clarify, a ViG backbone has a sequence of Grapher modules, which is a type of GNN layer (Grapher modules will be discussed in more detail in Section \ref{sec:grapher}). The graph of nodes in every Grapher module is constructed based on the similarity scores of the nodes, where each node picks the top-k most similar nodes as neighbors. The \textit{k}-NN graph construction has the following drawbacks: a node can have non-important nodes as neighbors, while missing out on important neighbors. Also, the graph construction is based on the similarity score calculation in the latent space of each layer, which might not result in the best graph topology to represent the input image.

To address the shortcomings of the \textit{k}-NN graph construction, SViG \citep{SViG} proposed a similarity-thresholded graph construction. This means that the normalized similarity score for each edge would be calculated, then the edges that have a normalized similarity score higher than a predefined threshold get picked. Nevertheless, this method of graph construction is still biased, since it depends on non-learnable hyper-parameters and similarity scores. 

In order to overcome the previously mentioned drawbacks of \textit{k}-NN and Similarity-Thresholded graph construction, we present the Learnable Reparameterized Graph Construction. There are two objectives to our methodology: \textit{(a)} To have learnable attention scores. \textit{(b)} To have a learnable threshold that can filter out irrelevant edges based on the attention scores. We borrow the concept of key-query dot product attention score calculation from Vision Transformers \citep{dosovitskiy2020vit}. In addition, we borrow the concept of Soft Threshold Reparametrization (STR) \citep{kusupati2020softthresholdweightreparameterization}. STR sparsified a network by removing unneeded weights that are lower than some threshold. To make the threshold learnable, they used the reparameterization trick \citep{reparamTrick}. We use a similar method to STR to be able to have learnable thresholds in LRGC that is going to be explained in detail in Section \ref{sec:lrgc_explanation}. \textbf{The contributions of ViG-LRGC are as follows:}
\begin{itemize}
    \item LRGC allows a node to \textit{attend} to other relevant nodes in a learnable fashion. 
    \item LRGC allows the threshold in each layer to be a learnable parameter that could be tuned on the data, rather than being a model hyper-parameter.
    \item We show empirically the pruning power of LRGC in different layers based on the data.
    \item ViG-LRGC outperforms the state-of-the-art GNN isotropic architectures in the literature that have a similar number of parameters.
\end{itemize}

In this paper, we first present a review of the literature in Section \ref{section:RelatedWork}. Then, a detailed explanation of our methodology is presented in Section \ref{sec:methodology}. That would be followed by presenting and discussing our results in Section \ref{sec:experiments}. Finally, we put forward a summary of our contributions and possible future work in Section \ref{sec:conclusion}.

\begin{figure}[h]
  \centering
  \includegraphics[width=0.65\linewidth]{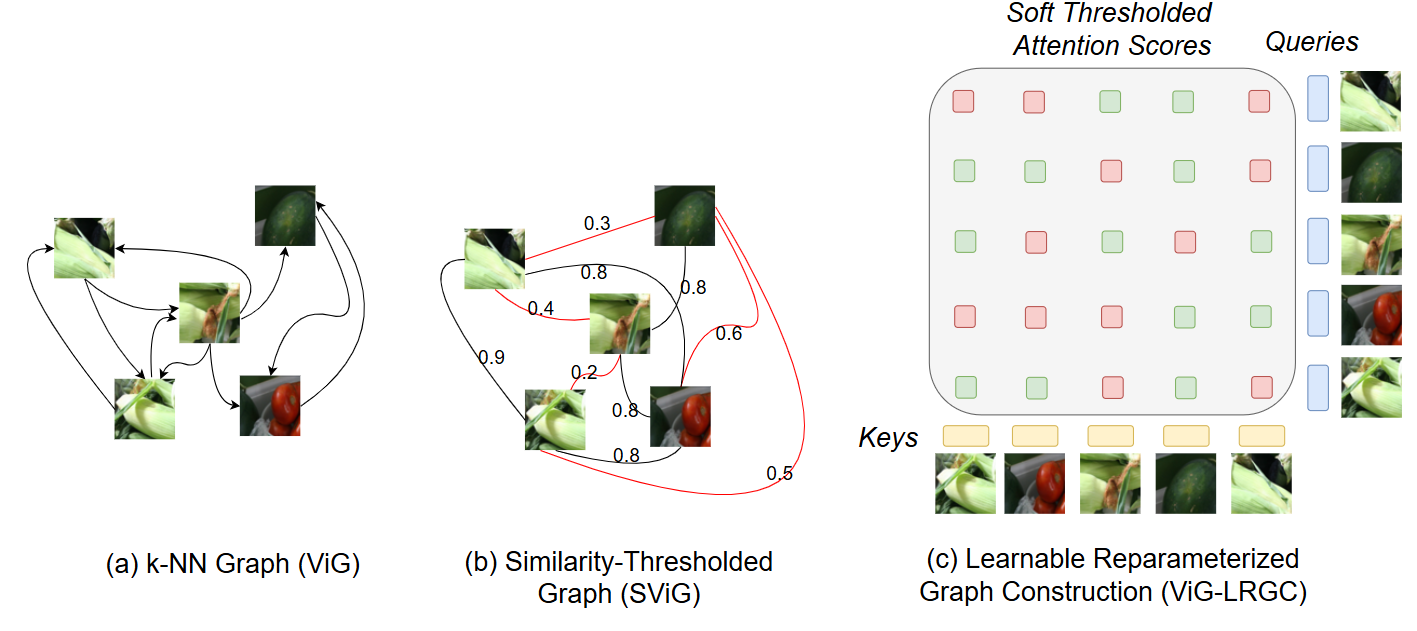}
  \caption{Graph Construction methodologies in ViG, SViG and ViG-LRGC. \textit{(a)} ViG uses \textit{k}-NN graph construction. In this example, k=2, thus each node has two incoming edges. \textit{(b)} SViG uses similarity-thresholded graph construction. The example shows hypothetical similarity scores on edges (ingoring self-edges), with a cut-off threshold of 0.7. \textit{(c)}ViG-LRGC \textit{(ours)} uses Learnable Reparameterized Graph Construction (LRGC). This example shows the calculation of key and query values of all nodes, followed by the attention score calculation. In this figure, red squares are the attention scores that are below the learnable threshold $\tau$ of that layer, whereas green squares represent attention scores are higher than the learnable threshold. In other words, green squares represent selected edges and red squares represent non-selected edges.
  }
  \label{fig:comparison}
\end{figure}

\section{Related Work}

\label{section:RelatedWork}

In the literature, there are two general directions of improvement to Vision Graph Neural Networks (ViG) \citep{han2022vig}, which are: performing an architectural search to find the best possible GNN architecture and using alternative graph construction methodologies. The architectural search task is related to finding the best Pyramid or hybrid GNN-CNN architecture, which we leave for future work. This section is going to focus mainly on the graph construction methodologies in the literature.

ViHGNN \citep{DBLP:conf/iccv/HanW00W23} introduced the concept of using a hyper-graph in GNN Vision models. A hyper-graph is a set of overlapping clusters of nodes. The clustering method they used was Fuzzy C-Means, which was an expensive operation to perform. After a hyper-graph is constructed, the there is an aggregation operation from all nodes in a cluster to a virtual super node, then another message passing  operation back to the nodes in the cluster.

There was another proposition in the literature to use axial graph construction, which was used by MobileViG \citep{DBLP:conf/cvpr/MunirAM23}, MobileViGv2 \citep{MobileViGv2_2024} and GreedyViG \citep{GreedyViG_CVPR_2024}. Axial Graph construction means that the receptive field of any patch is the set of patches on its row or column exclusively. MobileViG\citep{DBLP:conf/cvpr/MunirAM23} presented Dialted Axial Graph Construction, where an edge was constructed with every other patch on the axes. The major issue with this approach was aggregating noise from irrelevant neighbors. To solve this issue, GreedyViG\citep{GreedyViG_CVPR_2024} limited the receptive field of each node to nodes on the dilated axes with a Euclidean distance that is smaller than an estimate for $\mu - \sigma$. This filters our non-important neighbors for each node; however, important neighbors that are not on the dilated axes are dropped. Similar to GreedyViG, MobileViGv2 \citep{MobileViGv2_2024} used \textit{k}-NN graph construction on the dilated axes, where each node would pick the most similar k nodes on the dilated axes as neighbors. There is a significant drawback to this method, which is the fact that, similar to vanilla \textit{k}-NN, each node picks the same number of neighbors. This would entail aggregating from irrelevant neighbors and dropping important neighbors. 

To be able to have a more flexible graph representation, SViG \citep{SViG} proposed similarity thresholded graph construction. Similarity-thresholded graph construction assumed that the edge similarity scores followed a normal distribution. Then, a fixed threshold would be used, such that the edges with a normalized similarity score above that threshold will be selected. SViG also used a decreasing-threshold framework to avoid tuning each threshold in every layer independently. This approach, while efficient, makes an assumption that nodes can only attend to the most similar nodes, which might not always be the case. In addition, the decreasing threshold framework adds two hyper-parameters to the model, which need a grid search to find the best possible combination.

Apart from image classification on ImageNet-1k, Super Resolution (SR) was another task that was solved using GNN models. SR is the task where a low-resolution image is converted to a higher resolution image. The first attempt to tackle this problem with ViG-based models was Image Processing GNNs (IPG) \citep{Tian_2024_CVPR}. IPG introduces the concept of a local and a global neighborhood of a patch, where the local neighborhood represents the eight patches surrounding a patch and the global neighborhood represents the dilated grid of the patch grid. Each patch is downsampled, then upsampled again and the level of detail retained represents the patch importance, which is the then used to pick a \textit{k} value for that patch. Then, each patch selects the \textit{k}-NN from the local and global neighborhood, where each patch has its unique value for k. This approach has many disadvantages, which include the limited receptive fields of a node, the expensive downsampling and upsampling operation and finally, the fact that this graph construction method is tailored for SR.

In contrast to graph construction in ViG-based models, Vision Transformers (ViT) \citep{dosovitskiy2020vit} performed self-attention between all tokens in a sequence and thus constructed a fully connected graph. Some variations of ViT proposed other methods for graph construction between tokens in a sequence. For example, Swin Transformers \citep{liu2021swin} proposed the use of attention only within a predefined window, such that the size and aspect ratio of the window could change in different layers. Likewise, Shuffle Transformers \cite{huang2021shuffle} proposed performing self-attention within a predefined window, then shuffling the patches across different windows in following layers.

In our work, we propose a learnable method for graph construction that does not limit the receptive field of a node and can adapt the thresholds based on the data, instead of being fixed. We also replace similarity scores with learnable attention scores that can better represent the relationship between nodes.

\begin{figure*}[!ht]
  \centering
  \includegraphics[width=\textwidth]{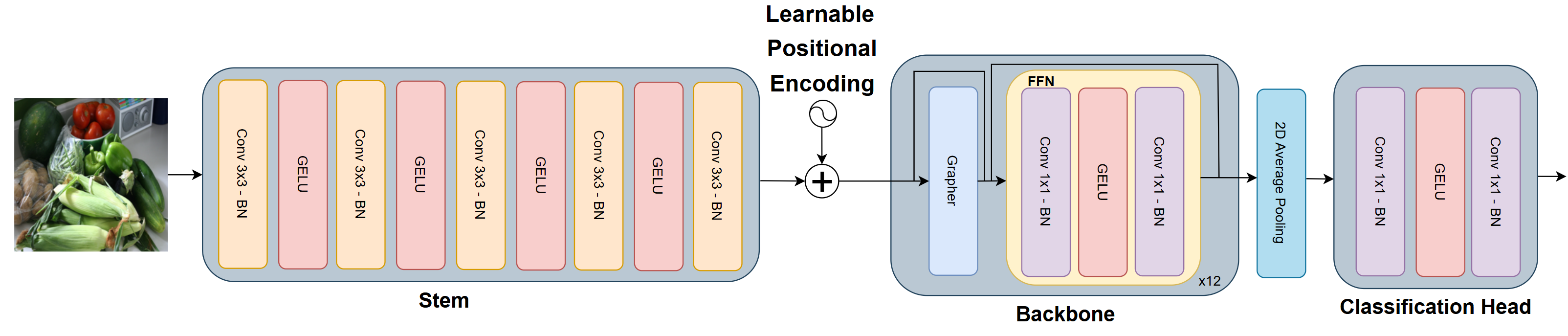}
  \caption{ViG-LRGC Model Architecture. The model has three main parts: a stem, a backbone and a classification head. The Stem is a CNN that acts as a feature extractor. That would be followed by adding learnable positional encoding. Each $1\times1\times D$ vector in the feature map represents a node in the graph. The next part of the model shows the backbone. The backbone is a sequence of 12 Grapher and FFN modules. Grapher modules are illustrated in more detail in Figure \ref{fig:maxrel}. After applying 2D Average pooling to the output of the backbone, the classification head is composed of two fully connected layers that outputs 1000 classification scores.}

  \label{fig:architecture}
\end{figure*}

\section{Methodology}\label{sec:methodology}
This section first presents the mathematical notation. Then, the overall architecture of the ViG-LRGC will be presented. Finally, the Learnable Reparameterized Graph Construction will be presented. 

\subsection{Notation}
The representation of an image is $I \in \mathbb{R}^{H \times W\times 3}$, such that $H$ is the height of the image and $W$ is the width of the image. An image is represented with a graph $G = (V,E)$, where $V$ is the set of vertices in graph $G$ and $E$ is the set of edges connecting these vertices. An directed edge from node $v_i$ to node $v_j$ is denoted by $e_{i,j} \in E$. The number of nodes in an image (or graph) is $N = |V|$. There are $L_G$ Grapher modules in the model, where each of them is denoted by $l$, such that $1 \leq l \leq L_G$. For each grapher module, the learnable threshold parameter is represented by $\tau_l$. In a Grapher module $l$, the feature vectors representing nodes of an image are denoted by $X^l = [x_1^l, x_2^l, x_3^l, ..., x_N^l]$, where a node $v_i$ in Grapher layer $l$ would be represented by a feature vector $x_i^l \in \mathbb{R}^D$, such that the embedding dimension of any node in any layer is $D$. For each pair of nodes $v_i$ and $v_j$ (or edge $e_{i,j}$), the attention score of these two nodes is $\alpha_{i,j}$.

\subsection{Overall ViG-LRGC Architecture}
ViG-LRGC architecture is adopted from ViG-Ti \citep{han2022vig}. The reason for the adoption of this model architecture is to have a fair comparison with other graph construction methodologies. The model is composed of a feature extractor (called a Stem), a backbone, and finally a classification head. The overall architecture of the model is shown in Figure \ref{fig:architecture}. 

The feature extractor is a shallow Convolutional Neural network (CNN). There are five CNN layers in the feature extractor, where the size of the output feature map is $\frac{H}{16}\times \frac{W}{16}\times D$. This is equivalent to $\frac{H}{16}\times \frac{W}{16}$ nodes with dimensions $1\times 1\times D$ each. It is important to note that each node represents a patch of size $19\times19\times 3$ in the input image. For each node, learnable positional embeddings of dimensions $1\times 1 \times D$ are added to preserve the positional information of each node.

The backbone is the main part of the model and is composed of a series of 12 sets of \textit{Grapher} and \textit{FFN} modules. A Grapher module is the type of GNN layer used. The architecture of the Grapher module will be explained in detail in Section \ref{sec:grapher}. The Feed Forward Network(FFN) module is a two-layer module that is there to increase the feature diversity and overcome oversmoothing. The FFN module is as follows:
\begin{equation}
    \label{eq:FFN}
    Y = GELU(XW_1)W_2 + X
\end{equation}
where $W_1$ and $W_2$ are learnable linear layers, $X$ is the output of the GNN and $Y \in \mathbb{R}^{N\times D}$ is the input to the next GNN layer. 

The backbone is followed by a 2D Average Pooling layer, which performs feature-wise averaging across all $N$ nodes in the graph. The output of the 2D Average Pooling layer is $1\times 1\times D$, which represents the whole image. This is followed by a classification head, which is composed of two fully connected layers. Since the problem under consideration is image classification, the output of the classification head is 1000 scores corresponding to the classes of ImageNet-1k.

For ViG-LRGC, the embedding dimension $D = 192$, which means that the feature vector of every node in any layer $x_i^l \in \mathbb{R}^{192}$. The input size is $224\times 224\times 3$, thus the dimension of the output feature map after the stem module is $14\times 14$, which means that a graph of an image has $196$ nodes in any layer (since we are using an isotropic architecture).

\subsection{Learnable Reparameterized Graph Construction}\label{sec:lrgc_explanation}
This section will first present the graph construction methodology; then the updated Grapher module architecture will be presented. Finally, a brief overview of the implementation will be provided.

\subsubsection{Graph Construction Methodology}
The main motivation of the Learnable Reparameterized Graph Construction (LRGC) methodology is to have a \textit{differentiable} hyper-parameter-free method for edge selection. Another motivation is to overcome the drawbacks in SViG \citep{SViG}; which are using similarity scores to select edges and the need for a hyper-parameter search to select the optimal threshold for each layer. We call the block that selects the edges an LRGC block. We assume that the LRGC block selects edges from a fully connected graph. Thus, we iterate over all possible (or candidate) edges in a fully connected graph. The purpose of the LRGC block is to assign a nonzero attention score to selected edges and a zero attention score for non-selected edges.

\begin{figure*}[!h]
  \centering
  \includegraphics[width=0.75\textwidth]{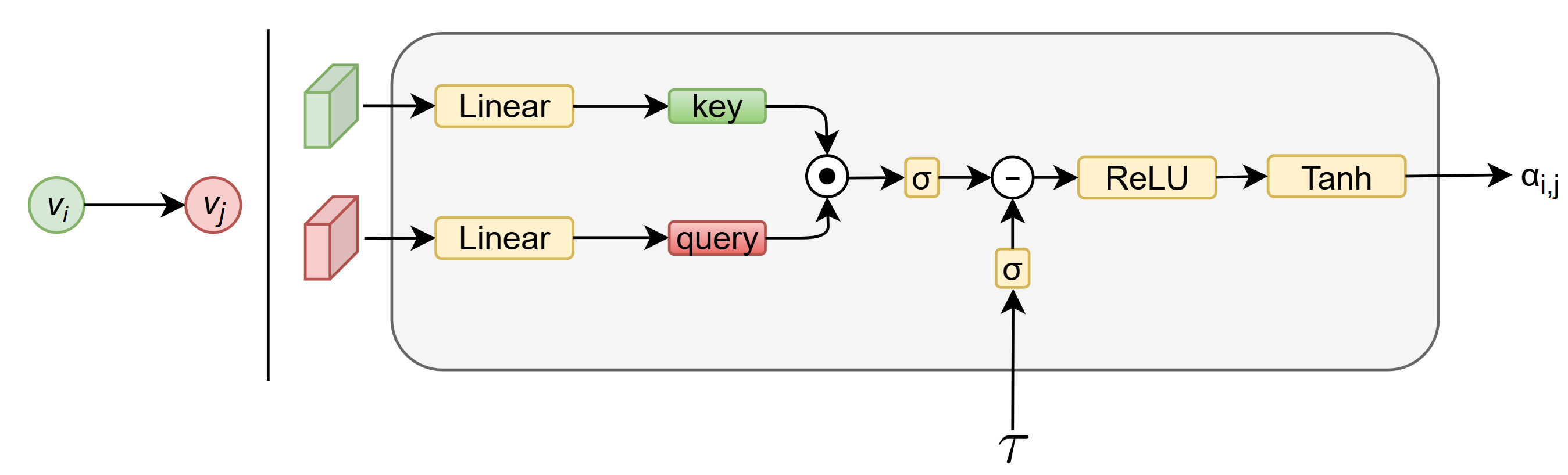}
  \caption{Learnable Reparameterized Graph Construction (LRGC) Block. This Figure corresponds to Equation \ref{eq:reparam_step3} \textit{Left:} There is an edge from node $v_i$ to $v_j$ under consideration. \textit{Right:} Nodes $v_j$ and $v_i$ are fed through two different linear layers to get their query and key respectively. We subtract the sigmoid of the learnable threshold $\tau$ from the sigmoid of the dot product of the key of node $v_i$ and query of node $v_j$. This subtraction is then passed through a ReLU to filter out edges with attention scores below $\tau$. The attention score is passed through a $\tanh$ to get $\alpha_{i,j}$, which is then used in the updated Grapher module (shown in Figure \ref{fig:maxrel}). 
  }
  \label{fig:lrgc}
\end{figure*}

LRGC is based on concepts of Self-Attention in Vision Transformers (ViT) \citep{dosovitskiy2020vit} and Soft Threshold Reparameterization (STR) \citep{DBLP:journals/corr/abs-2002-03231}. Figure \ref{fig:lrgc} shows the LRGC methodology. In that figure, a candidate edge from node $v_i$ to node $v_j$ is shown. In addition, the \textit{learnable} threshold $\tau$ is shown as an input to the block. This procedure is performed for all possible edges in the graph. It is important to note that each grapher module has its own $\tau$.

Similar to the self-attention of visual tokens in Vision Transformers \citep{dosovitskiy2020vit}, we calculate the attention score between every pair of nodes. For nodes $v_i$ and $v_j$, the attention score $\alpha^{'}_{i,j}$ is calculated as follows:
\begin{equation}
\label{eq:reparam_step1}
\begin{split}
    &key(v_i) = W_{key} h_i \\
    &query(v_j) = W_{query}h_j\\
    &\alpha^{'}_{i,j} = key(v_i)\cdot query(v_j) \\
\end{split}
\end{equation}
where $W_{key}$ and $W_{query}$ are learnable linear layers. A \textit{key} and a \textit{query} of a node are linear projections of the hidden representation of that node. The dot product of the key of node $v_i$ and the query of node $v_j$ represents a lookup in a query table or the importance of a node relative to another.

We borrow the concept of soft threshold reparameterization from STR \citep{DBLP:journals/corr/abs-2002-03231}. We use the attention score $\alpha^{'}_{i,j}$ calculated in Equation \ref{eq:reparam_step1} as follows:
\begin{equation}
\label{eq:reparam_step2}
\begin{split}
    &\alpha^{''}_{i,j} = ReLU(\sigma(\alpha^{'}_{i,j}) - \sigma(\tau))\\
    &\alpha_{i,j} = \tanh(\alpha^{''}_{i,j})
\end{split}
\end{equation}
We subtract the sigmoid of the learnable threshold $\tau$ from the sigmoid of $\alpha^{'}_{i,j}$ and apply a $ReLU$ to the difference. The intuition is to have a zero score when $\alpha^{'}_{i,j}\leq \tau$ and a nonzero score otherwise. The purpose of using sigmoids with the threshold $\tau$ and $\alpha^{'}_{i,j}$ is to make training numerically stable, since the difference will always be smaller than 1. Finally, $\tanh$ is applied as a scaling function that preserves the value of zero attention scores.

Putting together Equations \ref{eq:reparam_step1} and \ref{eq:reparam_step2}, we calculate the attention scores as follows:
\begin{equation}
\label{eq:reparam_step3}
\begin{split}
    &\alpha_{i,j} = \tanh(ReLU(\sigma( W_{key} h_i \cdot W_{query}h_j)  - \sigma(\tau)))\\
\end{split}
\end{equation}

It is important to mention that the purpose of this methodology is to prune non-needed edges. Thus $\tau$ should be initialized with a low value to be able to have a fully connected graph during the first optimization step, and then allow the model to prune edges that are not important. We choose to initialize $\tau$ at -1 for all layers. We show empirically in Figure \ref{fig:avg_no_nodes} that this initialization value allows the graph to be fully connected in the first optimization step.
 
\subsubsection{Updated Grapher Module} \label{sec:grapher}
We adopt the Grapher module architecture from ViG-Ti \citep{han2022vig} with only one modification: we add the learned attention score to the Max-Relative Graph Convolution. The updated Grapher module can be described via the following set of equations:
\begin{equation}
\label{eq:Grapher}
\begin{split}
   &{x_j}^{'} = {x_j}^{l}W_{in}\\
   &{x_j}^{''} = \max(\alpha_{i,j}({{x_i}^{'} - {x_j}^{'}) | v_i \in \textit{Neighborhood}({v_j})}) \\
   &{x_j}^{'''} = GELU([{x_j}^{'},{x_j}^{''}]W_{update})W_{out} \\
   &{x_j}^{l+1} =  {x_j}^{'''} + x_j^{l} \\
\end{split}
\end{equation}
where $W_{in}$, $W_{update}$ and $W_{out}$ are learnable linear layers. In addition, $W_{update}$ is a multi-head update layer with 4 update heads. $\alpha_{i,j}$ is the attention score that was obtained from Equation \ref{eq:reparam_step3}. $x_j^{'}$, $x_j^{''}$ and $x_j^{'''}$ are intermediate representations of node $v_j$ in the Grapher module.

\begin{figure*}[]
  \centering
  \includegraphics[width=\textwidth]{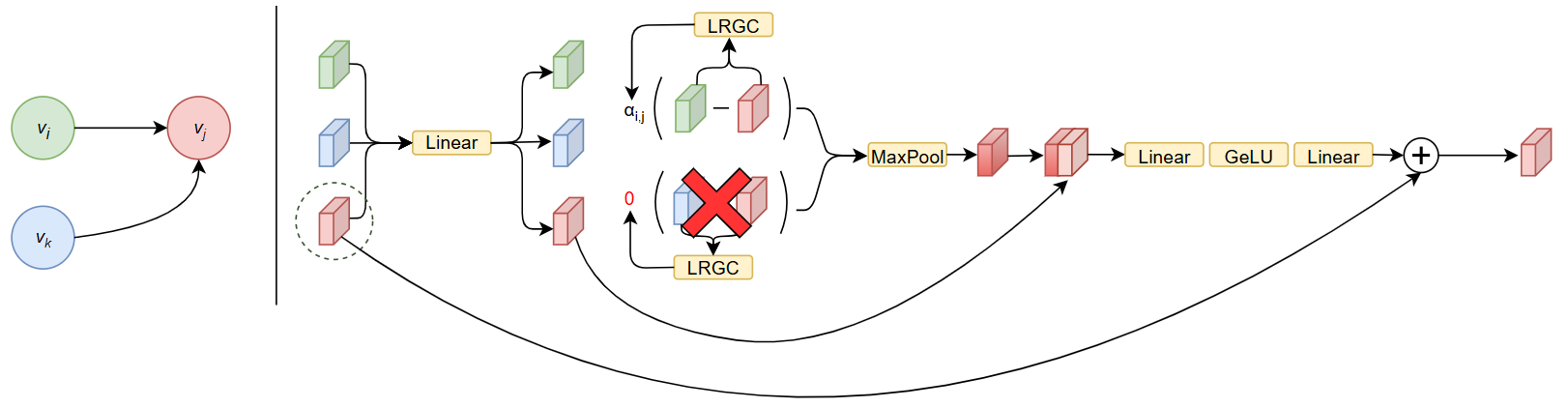}
  \caption{Updated Grapher module. This Figure corresponds to Equation \ref{eq:Grapher}. \textit{Left:} This is a hypothetical graph with three nodes: $v_i$, $v_k$ and $v_j$, where there is an edge from node $v_i$ to node $v_j$ and an edge from node $v_k$ to node $v_j$. The target node is node $v_j$. \textit{Right:} The inputs are $x_i, x_j$ and $x_k$, which are the representations of nodes $v_i$, $v_j$ and $v_k$ (with the target node representation $x_j$ being circled). All nodes are fed through a linear layer. Then, for every edge, both nodes are fed to an LRGC module to calculate the attention scores $\alpha_{i,j}$ and $\alpha_{k,j}$. In this example, $\alpha_{k,j} = 0$, which means that the edge was not selected. The residuals, which are multiplied by the attention scores $\alpha_{i,j}$ and $\alpha_{k,j}$, are fed to a Max-Pooling layer. This is followed by a concatenation with the representation of the target node, then two linear layers and a residual connection. 
  }

  \label{fig:maxrel}
\end{figure*}

Figure \ref{fig:maxrel} shows the updated Grapher module. In that figure, $v_i$ and $v_k$ are the source nodes; while $v_j$ is the target node. This means that there are two edges to consider: an edge from node $v_i$ to node $v_j$ and another one from node $v_k$ to node $v_j$. First, the node feature vectors are fed to a linear layer. Then, every pair of nodes is fed into the LRGC block to get an attention score for the edge being considered. That would be followed by multiplying the attention score by the subtraction of the feature vectors of the source and target nodes. This multiplication operation with $\alpha_{i,j}$ is important to allow the model to be differentiable. Then, the residuals that were multiplied by the attention scores go through a feature-wise max-pooling operation. Effectively, non-selected edges would have the residuals multiplied by zero, which means that the feature values do not affect the max-pooling operation. 

This would be followed by a concatenation with the feature vector of the target node, then a Linear, a GELU \citep{gelu} activation function and a Linear layer. Finally, there is a residual connection with the input feature vector of the target node. 

\subsection{LRGC Implementation} \label{sec:lgrc_impl}
We use the flexible aggregation from SViG \citep{SViG} because it does not place a constraint on each node to have a specific number of neighbors in the graph. However, there is a need to start with a fully connected graph to allow the model to prune the edges in a fair manner. This means that $\tau$ is initialized with a low value and that all edges are selected during the first optimization step. This places a limitation on the computational power needed for the implementation of ViG-LRGC, which in turn affects the batch size (discussed in more detail in Section \ref{sec:exp_setup}).

\section{Experiments} \label{sec:experiments}
In this section, the dataset used is described. Then, the experimental setup used for our experimentation is presented. We then compare our model with the state-of-the-art models in the literature. Finally, we perform an analysis for our ViG-LRGC model. 

\subsection{Dataset}
We used ImageNet-1k (ImageNet ILSVRC 2012) dataset \citep{ImageNet} in our experiments. This dataset could be obtained via a license from \url{https://www.image-net.org/download}. There are 1.2 million images in the training set and 50k images in the testing set. This dataset is used for the image classification task and it has 1000 classes.

\begin{table}[b]
  \caption{Hyper-parameters used in the training of ViG-LRGC}
  \label{tab:hyperparams}
  \begin{adjustbox}{width=0.7\columnwidth,center}
  \begin{tabular}{c|c}
    \toprule
    Hyper-parameter&Value set for ViG-LRGC\\
    \midrule
    Epochs& 300 \\
    Optimizer& AdamW \citep{adamw}\\
    Batch size & 256\\
    Start learning rate (LR) & 2e-3\\
    Learning rate Schedule & Cosine\\
    Warmup epochs & 20\\
    Weight decay & 0.05\\
    Label smoothing\citep{GoogleNetv3} & 0.1\\
    Stochastic path\citep{huang2016deep} & 0.1\\
    Repeated augment\citep{hoffer2020augment} & \checkmark \\
    RandAugment \citep{randaugment} & \checkmark \\
    Mixup prob. \citep{mixup} & 0.8 \\
    Cutmix prob. \citep{cutmix} & 1.0 \\
    Random erasing prob. \citep{erasing} & 0.25 \\
    Exponential moving average & 0.99996 \\
  \bottomrule
\end{tabular}
\end{adjustbox}
\end{table}
\begin{table*}[!b]
  \caption{Comparing ViG-LRGC to GNN architectures. This table shows the results of the experiments that we ran using a batch size of 256. It can be observed that ViG-LRGC surpasses the ViG-Ti baseline by 3.8\% on Top-1 Accuracy while surpassing SViG-Ti by 2.4\%}
  \label{tab:comparison}
  \begin{adjustbox}{width=0.6\textwidth,center}
  \begin{tabular}{c|c|c|c|c}
    \toprule
    Model&Params (M)&FLOPs (B)&Top-1&Top-5\\
    \midrule
    \textbf{ViG-LRGC(ours)}&\textbf{8.1}&\textbf{1.5}&\textbf{74.3}&\textbf{92.3}\\
    ViG-Ti\citep{han2022vig}&7.2&1.3&70.5&89.9\\
    SViG-Ti\citep{SViG}&7.2&1.3&71.9&90.8\\

  \bottomrule
\end{tabular}
\end{adjustbox}
\end{table*}

\subsection{Experimental Setup}\label{sec:exp_setup}
In our implementation, we used PyTorch 1.7.1 \citep{pytorch}, PyTorch Geometric (PyG) 2.3.1 \citep{Fey/Lenssen/2019} and PyTorch Image Models (Timm) 0.3.2 \citep{rw2019timm}. We use PyTorch Distributed Data Parallel (DDP) to train our model on a DGX-1 server with 8 NVIDIA V100 GPUs (with 32 GB of memory each). During training, the same batch size should be used for all optimization steps. However, due to the need to have a fully connected graph in the first optimization step (as described in Section \ref{sec:lgrc_impl}), the effective batch size used is 256 due to computational power limitations. For a fair comparison, we use the same training hyper-parameters used for ViG \citep{han2022vig}. The hyper-parameters used for training are shown in Table \ref{tab:hyperparams}. For each reported result, we train the model from scratch on ImageNet for 300 epochs with 20 warmup epochs and 10 cooldown epochs. We resize all images to $224\times 224\times 3$. The optimizer we used was AdamW \citep{adamw}. The learning rate schedule used was a Cosine Annealing schedule with a base learning rate of $2e-3$. The data augmentation methods used were CutMix \citep{cutmix}, MixUp \citep{mixup}, RandAugment \citep{randaugment}, repeated augment \citep{hoffer2020augment} and random erasing \citep{erasing}. For all layers, $\tau$ is initialized with -1.

\subsection{Experimental Results}
In order to have a fair comparison, we focus on state-of-the-art isotropic models that have a similar number of parameters and that focus solely on graph construction. We choose to have a comparison with ViG-Ti \citep{han2022vig} baseline and SViG-Ti \citep{SViG}, which adopt the same model architecture as ViG-Ti. 

Since we use a batch size of 256 to train ViG-LRGC, we replicate the training experiments for both ViG-Ti \citep{han2022vig} and SViG-Ti \citep{SViG} with the same batch size to have a fair comparison. We present our experimental results in Table \ref{tab:comparison}.

\subsubsection{Effect of the batch size on the training}
For the same model, we observe that using a smaller batch size yields lower results (while still using the same hyper-parameters and training setup). For ViG-Ti, using a batch size of 256 yields a Top-1 Accuracy of 70.5\%, opposed to a Top-1 of 73.9\% when using a batch size of 1024. Similarly, the Top-1 Accuracy of SViG-Ti, when using a batch size of 256, is 71.9\%, while the Top-1 Accuracy of SViG-Ti becomes 74.6\% when using a batch size of 1024.

\subsubsection{Comparison to the State Of The Art}
The Top-1 Accuracy of ViG-LRGC is 74.3\% and the Top-5 Accuracy is 92.3\%. Using the same batch size of 256, we show that our ViG-LRGC model surpasses the baseline ViG-Ti with 3.8\% on Top-1 Accuracy and 2.4\% on Top-5 Accuracy. Our ViG-LRGC also surpasses SViG-Ti with 2.4\% on Top-1 Accuracy and 1.5\% for Top-5 Accuracy. Our ViG-LRGC model outperforms the state-of-the-art GNN isotropic models, that are trained with the same hyper-parameters.

\subsection{Computational Complexity of LRGC}
In both ViG and SViG, the computational complexity of the graph aggregation was dominated by the calculation of the similarity matrix. The similarity score calculation computational complexity is $O(N^2*D)$, where $D$ is the embedding size and $N$ is the number of nodes in an image \citep{SViG} Assuming that $N \approx D$, then the complexity becomes $O(N^3)$.

For ViG-LRGC, the multiplication of a feature vector of a node by the weights of $W_{key}$ or $W_{query}$ has a computational complexity of $O(D^2)$, since the size of both the input and output feature vectors is $1\times\ 1\times D$. To calculate the values of the keys and the queries of all $N$ nodes in an image, the complexity becomes $O(N*D^2)$. The computational complexity of the dot product of the key and query is $O(D)$. Since we assume a fully connected graph, the calculation of all attention scores has a computational complexity of $O(D*N^2)$ since there are $N^2$ edges in a fully-connected graph. If we assume that $N \approx D$, then the total complexity becomes $O(N*D^2 + D*N^2) \approx O(N^3)$. Thus, the complexity of ViG-LRGC is similar to ViG-Ti and SViG-Ti. This is experimentally proven by the fact that ViG-LRGC has a FLOPs count of 1.6B, compared to 1.3B of both ViG-Ti and SViG-Ti. While the number of parameters of a ViG-LRGC model is slightly larger than that of ViG-Ti and SViG-Ti, the slight increase is justifiable due to the increase in performance.

\begin{figure}
  \centering
  \includegraphics[width=0.7\columnwidth]{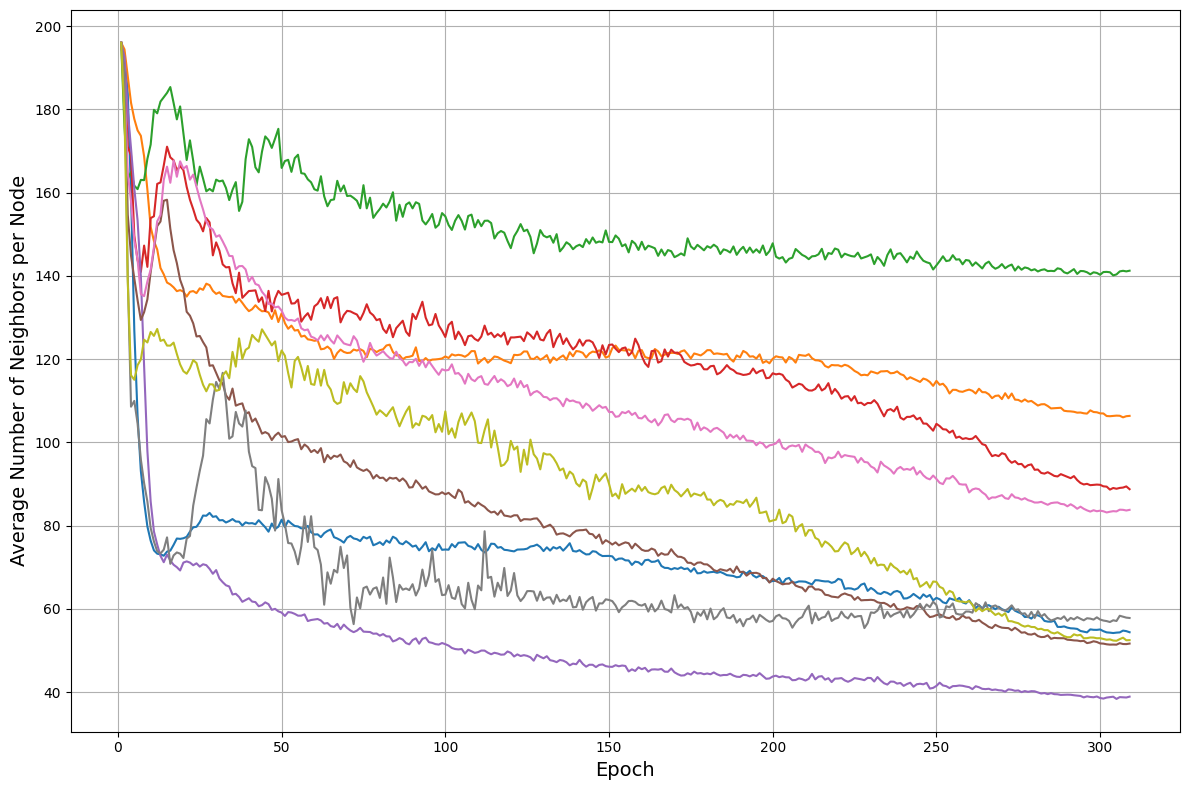
  }
  \caption{This figure shows the average number of neighbors per node for the training epochs. Each line represents the average number
of neighbors per node for a different layer. This Figure proves the pruning power of LRGC for non-needed edges.
  }

  \label{fig:avg_no_nodes}
\end{figure}
\subsection{Empirical Proof for the Pruning Power of LRGC}
To prove the pruning power of ViG-LRGC, Figure \ref{fig:avg_no_nodes} shows the average number of neighbors per node for some layers. This figure shows how the average number of neighbors per node changes across the training epochs. The average number of neighbors can be seen as a reflection of the change in the threshold $\tau$ in each layer. As seen, the average number of nodes starts from the maximum possible value (196). During the first 50 epochs, the average number of nodes first decreases, then increases slightly then drops again. Starting from epoch 50, the average number of nodes decreases steadily. Each layer tunes the learnable threshold $\tau$ based on the training data. Opposed to hard-thresholding in SViG\citep{SViG}, the soft-thresholding allows the learning of the threshold $\tau$. This is similar to the trend of the weight pruning that is observed in STR \citep{kusupati2020softthresholdweightreparameterization}.

\section{Conclusion}\label{sec:conclusion}
In conclusion, we present a novel method for graph construction in Vision Graph Neural Networks. Our Learnable Reparameterized Graph Construction (LRGC) methodology proposes a powerful alternative for the similarity score calculation through the learnable attention scores. Moreover, LRGC presents learnable thresholding by utilizing the reparameterization trick to learn the threshold for each layer; thus, removing the need for a grid search over hyper-parameters. We prove that LRGC has a great pruning power for non-needed edges. Our methodology is superior to the state-of-the-art GNN models with similar sizes, with a minimal increase in FLOPs. We hope that this work inspires future work on Pyramid or GNN-CNN hybrid models that use LRGC as a learnable method for graph construction.

\bibliographystyle{unsrtnat}
\bibliography{main}  

\end{document}